\documentclass[conference]{IEEEtran}
\IEEEoverridecommandlockouts
\usepackage{cite}
\usepackage{amsmath,amssymb,amsfonts}
\usepackage{algorithmic}
\usepackage{graphicx}
\usepackage{textcomp}
\usepackage{xcolor}
\usepackage{url}
\usepackage{lipsum}
\usepackage{comment}
\def\BibTeX{{\rm B\kern-.05em{\sc i\kern-.025em b}\kern-.08em
    T\kern-.1667em\lower.7ex\hbox{E}\kern-.125emX}}

\begin{document}

\title{Efficient Causal Discovery for\\Robotics Applications\\
%\title{Paper Title*\\
\thanks{
\textsuperscript{1}School of Computer Science, University of Lincoln, UK.\newline
\textsuperscript{2}Dept. of Information Engineering, University of Padua, Italy.\newline
This work has received funding from the EU H2020 research \& innovation programme -- grant agreement 101017274 (DARKO).}
}

%\author{\IEEEauthorblockN{Luca Castri}
%\IEEEauthorblockA{%\textit{L-CAS} \\
%\textit{University of Lincoln}\\
%Lincoln, UK \\
%lcastri@lincoln.ac.uk}
%\and
%\IEEEauthorblockN{Sariah Mghames}
%\IEEEauthorblockA{%\textit{L-CAS} \\
%\textit{University of Lincoln}\\
%Lincoln, UK \\
%smghames@lincoln.ac.uk}
%\and
%\IEEEauthorblockN{Nicola Bellotto}
%\IEEEauthorblockA{%\textit{Department of Information Engineering} \\
%\textit{University of Padua}\\
%Padua, Italy \\
%nbellotto@dei.unipd.it}
%}
\author{Luca Castri\textsuperscript{1}, Sariah Mghames\textsuperscript{1} and Nicola Bellotto\textsuperscript{1,2}}

\maketitle

\begin{abstract}
Using robots for automating tasks in environments shared with humans, such as warehouses, shopping centres, or hospitals, requires these robots to comprehend the fundamental physical interactions among nearby agents and objects. Specifically, creating models to represent cause-and-effect relationships among these elements can aid in predicting unforeseen human behaviours and anticipate the outcome of particular robot actions. To be suitable for robots, causal analysis must be both fast and accurate, meeting real-time demands and the limited computational resources typical in most robotics applications.
In this paper, we present a practical demonstration of our approach for fast and accurate causal analysis, known as Filtered~PCMCI~(F-PCMCI), along with a real-world robotics application. The provided application illustrates how our F-PCMCI can accurately and promptly reconstruct the causal model of a human-robot interaction scenario, which can then be leveraged to enhance the quality of the interaction.
%In this paper, we present a practical view of our approach for fast and accurate causal analysis named Filtered PCMCI (F-PCMCI) and a real-world robotics application of it. The proposed application shows how our F-PCMCI is able to reconstructs the causal model of a human-robot interaction scenario in a fast and accurate way and the obtained causal model can be further exploited to enhance the quality of the interaction.
\end{abstract}


\section{Introduction}
The increased use of robots in numerous sectors, such as industrial, agriculture and healthcare, represents a turning point for their progress and growth. However, it requires also new approaches to study and design effective human-robot interactions. 
A robot, sharing the working area with humans, must accomplish its task taking into account that its actions may lead to unpredicted responses by the individuals around it, while at the same time
taking into account the execution
time and the computational cost for
completing the task.
Rapid and accurate understanding of the cause-and-effect relationships in the environment will allow the robot to reason on its own actions. The latter represents a crucial step towards effective human-robot interactions and collaborations.

\begin{figure}[t]\centering
\includegraphics[trim={0cm 0cm 4.5cm 4cm}, clip, width=0.93\columnwidth]{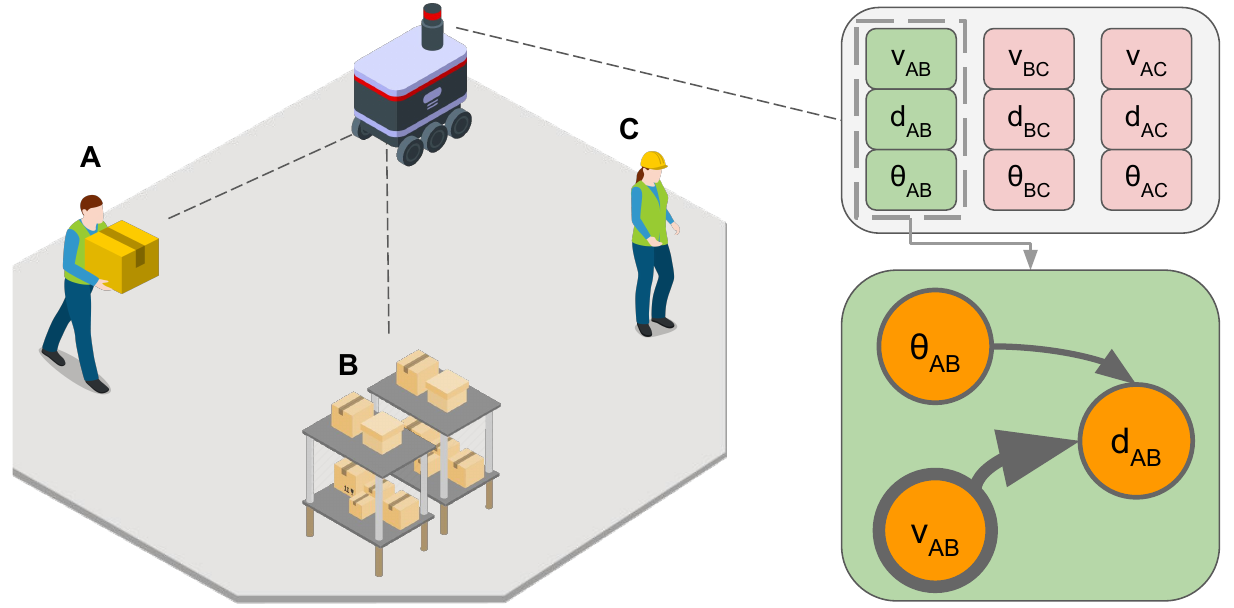}
\caption{
A mobile robot in a warehouse-like settings observes the interaction between agents A and B. With our method, the robot can ignore interactions AC and BC, since agent B is a static object and agent C is a stationary human not participating in the interaction.}
\label{fig:intro}
\end{figure}

% Literature review
Causal inference~\cite{Pearl2009} is an active research area in various fields, including robotics~\cite{hellstrom2021relevance}, human-human and human-robot spatial interactions~(HRSI)~\cite{castri2022causal,castri2023enhancing}. Over the past decades, numerous causal discovery methods have been developed for static and time-series data\cite{glymour_review_2019,runge_detecting_2019}.
% Motivation
However, most of these works overlooked a feature that is crucial for real-world applications, i.e. the computational cost of the causal analysis when applied to scenario with limited hardware resources and real-time requirements, such as autonomous robotics. Indeed, causal analysis of complex and dynamical systems is extremely demanding in terms of time and hardware resources~\cite{runge_detecting_2019}, making it a challenge for autonomous robotics~\cite{castri2023continual}.
This paper presents an overview of our approach Filtered PCMCI~(F-PCMCI) along with a real-world robotics application. 
F-PCMCI offers an all-in-one solution that identifies the causal features characterising the system and builds a causal model directly from time-series data. As a result, the causal discovery process turns out faster and more accurate, rendering it well-suited for applications in autonomous robotics.
% Example
In an automated warehouse scenario (see Fig.~\ref{fig:intro}), where a robot observes interactions among objects and humans (e.g., workers and shelves), identifying relevant features from the robot's sensors is vital. For instance, features like human-shelf distance/angle and human velocity are significant for describing observed interactions, while others (e.g., unrelated humans) can be ignored. F-PCMCI allows the robot to exclude unnecessary features, constructing a causal model solely from the crucial ones. This causal model can then be used to better understand and predict the interaction of the involved agents.

To summarise, our approach F-PCMCI is an effective algorithmic solution to select the most meaningful features from a set of variables and build a causal model from such selection. To this end, we significantly enhance speed and accuracy of the causal discovery making it suitable for robotics applications. 

\section{Filtered-based Causal Discovery}
\begin{figure*}[t]\centering
\includegraphics[trim={0cm 0cm 0cm 9.3cm}, clip, width=\textwidth]{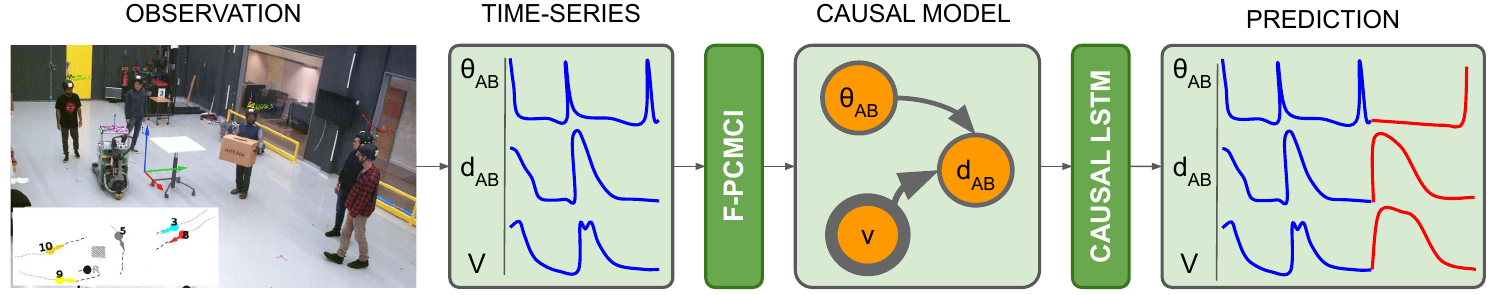}
\caption{Pipeline showing the use of F-PCMCI for modeling and predicting spatial interactions in a robot's intralogistics scenario~\cite{castri2023enhancing}.}
\label{fig:fpcmci_flowchart}
\end{figure*}
Our Filtered PCMCI~(F-PCMCI)~\cite{castri2023enhancing} extends one of the state-of-the-art causal discovery methods, i.e. PCMCI~\cite{runge_detecting_2019}, augmenting it with a feature-selection algorithm that is able to identify the correct subset of variables for causal analysis, starting from a predefined set of them. 
It comprises two main components: the feature selection step and the causal discovery process.
In the feature selection step, we use a Transfer Entropy (TE)-based method to ``filter'' important features and identify potential associations from a larger set of variables. 
This filter calculates TE between each variable and the others. If an association is found, the variables are considered in subsequent causal discovery, performed by the PCMCI, to assess their causal relationship~\cite{castri2023enhancing}. The TE filter therefore provides a set of variables and a hypothetical causal model which contribute to speeding up and improving the accuracy of the causal discovery.
%In practice, for each variable in our predefined set, the filter computes the TE from every other variable in the same set. If an association between two variables exists, then they are correlated and can be considered in the following causal discovery step to determine whether and how they are causally linked~\cite{castri2023enhancing}. The TE filter therefore provides a set of variables and a hypothetical causal model, further validated through a PCMCI causal analysis. The combination of the two results contributes to speeding up and improving the accuracy of the causal discovery.

To aid the evaluation and application of our method on real-world problems, we have implemented a Python package of F-PCMCI, which is publicly available on GitHub\footnote{https://github.com/lcastri/fpcmci} and PyPI\footnote{https://pypi.org/project/fpcmci}. Additionally, we have provided a comprehensive web page and various tutorials that explain how to use F-PCMCI, starting from input time-series data to the output causal model.

\section{Robot Application}
We applied our approach to model and predict spatial interactions in a human-robot collaboration scenario for intralogistics~\cite{castri2023enhancing}, as part of a large EU project (DARKO\footnote{https://cordis.europa.eu/project/id/101017274}). As shown in Fig.~\ref{fig:fpcmci_flowchart}, this involves three main steps:
(\textit{i})~extracting time-series of sensor data (i.e. trajectories) from human spatial interactions; 
(\textit{ii})~reconstructing the causal model using F-PCMCI; 
(\textit{iii})~embedding the causal model in a neural network, LSTM-based prediction system. 
We extracted human and robot trajectories from the TH{\"O}R dataset~\cite{thorDataset2019}, which uses helmets and infrared cameras to track agents' motion. The dataset features a shared working area comprising people, target positions, static objects, and a moving robot in a warehouse-like environment. People carry boxes and navigate the environment alone or in groups, simulating intralogistics activities. The robot, a small autonomous forklift, moves in a safe but socially unaware manner, following a predefined path with a maximum speed of 0.34 m/s. It projects its current motion intent onto the floor in front of it using a mounted beamer. We chose this dataset for its diverse range of interactions, including humans, robot, and static objects~(Fig.~\ref{fig:fpcmci_flowchart}, left image).
Our LSTM-based prediction system includes an encoder-decoder network with an input-attention layer, which selects the most relevant time steps from the observation window, and a self-attention one to identify the key prediction cues. The latter integrates also the causal discovery output as a non-trainable parameter.

A detailed explanation of the system and experimental results can be found in our recent paper~\cite{castri2023enhancing}. It compares our F-PCMCI to the standard PCMCI in terms of accuracy and execution time, outperforming the latter in both metrics. It also shows how F-PCMCI's causal model improves the accuracy of the spatial interaction prediction, for a possible robot's intralogistics scenario, compared to a non-causally informed system. Further details about the Python implementation of the prediction system are provided in the companion web page\footnote{https://github.com/lcastri/cmm\_ts}.

\section{Conclusion}
In this paper, we motivated the need and provided an overview of an efficient causal discovery algorithm for robotics applications. By integrating a feature-selection module based on transfer entropy, our F-PCMCI enhances the accuracy and computational efficiency of causal discovery, making it well-suited for autonomous robot applications. In particular, the paper shows a robot-assisted intralogistics scenario in which the F-PCMCI causal model improves the quality of the prediction of spatial interactions in a warehouse-like environment.

Inspired by recent research on continual learning and causality~\cite{castri2023continual}, our future work aims to enhance F-PCMCI with a real-time strategy for adaptive causal modeling. This would empower the robot to continually improve the causal model and adapt to changing scenarios as new data arrives.

\bibliographystyle{IEEEtran}
\bibliography{IEEEabrv,references.bib}
\end{document}